\documentclass[runningheads,a4paper]{llncs}

\usepackage{lscape}
\usepackage{subfigure}
\usepackage{amssymb}
\usepackage{amstext}
\usepackage{amsmath}
\usepackage{multirow}
\usepackage{amssymb}
\usepackage[noend]{algorithmic}
\usepackage{latexsym}
\usepackage{graphicx}

\usepackage{verbatim}

\begin{document}
\title{Analysing the behaviour of robot teams through relational sequential pattern mining}

\author{Grazia Bombini${^1}$, Raquel Ros${^2}$, Stefano Ferilli${^1}$, Ramon Lopez de Mantaras${^3}$ } 
\institute{
${^1}$University of Bari ``Aldo Moro'', Department of Computer Science,
70125 Bari, Italy\\
\{gbombini, ferilli\}{@}di.uniba.it\\
${^2}$Department of Electrical and Electronic Engineering,
Imperial College, UK \\
r.ros-espinoza@imperial.ac.uk\\
${^3}$IIIA - Artificial Intelligence Research Institute, CSIC - Spanish Council for Scientific Research,
Campus UAB, 08193 Bellaterra, Spain\\
mantaras@iiia.csic.es
}

\authorrunning{G. Bombini et al.}
\titlerunning{Analysing the behaviour of robot teams}

\maketitle

\begin{abstract}
{ 
This report outlines the use of  a relational representation in a Multi-Agent domain 
to model the behaviour of the whole system.
A desired property in this systems is the ability of the team members to work together
to achieve a common goal in a cooperative manner.
The aim is to define a systematic method to verify the effective collaboration 
among the members of a team
and comparing the different multi-agent behaviours.
Using external observations of a Multi-Agent System 
to analyse, model, recognize agent behaviour could be very useful to direct team actions.

In particular, this report focuses on the challenge of autonomous unsupervised
sequential learning of the  team's behaviour from observations. 
Our approach allows  to learn a symbolic sequence (a relational representation) to translate 
raw multi-agent, multi-variate observations of a dynamic, complex environment, into a set of 
sequential behaviours that are characteristic of the team in question, represented by a set
of sequences expressed in first-order logic atoms.
We propose to use a relational learning algorithm to mine meaningful frequent patterns among the 
relational sequences to characterise team behaviours.

We compared the performance of two teams in the RoboCup four-legged league environment, 
that have a very different approach to the game. One uses a Case Based Reasoning approach,  the other 
uses a pure reactive behaviour.

%\keywords{}
}
\end{abstract}

\section{Introduction}
\label{sec:introduction}

Action selection in robotics is considerate a challenging task in different fields of 
Artificial Intelligence.
An autonomous robot has to reason about the state of the environment and rationally act 
in order to complete a given task. 

The complexity of each individual ability, and therefore the overall robot’s behaviour
design, is related to the complexity of the environment where the robot 
carries out the task: the higher the complexity of the environment, the more
challenging the robot’s behaviour design. 
Indeed, in real world unpredictable situations always occur, therefore is not possible to design completely controllable 
environments. Besides,  creating highly controlled scenarios to  decrease the difficulty of the task, 
makes it less realistic.

A robot needs multiple capabilities to perform a task, often divided into subtasks.
For instance, to  move a ball towards a goal point it should possess skills such as
object detection, perception of the environment, building of an internal world
model, making decisions when planning the task, navigation while avoiding
obstacles, execution of planned actions, and recovering from failure.
Thus, the reasoning engine must be capable of dealing with high uncertainty in the robot’s perception 
(incoming information of the world), and be robust in case of failure, since the outcomes of the actions
performed are unpredictable. 
In this kind of environments, dynamic and unpredictable, the agent must be able to detect if the 
selected actions for a given state of the environment are still applicable when the state evolves.
If they are, then the agent continues with the initial plan. Otherwise, it must either correct
the selected actions or re-plan.

Moreover, in the case of a robot team, they have to  jointly execute the selected actions, 
and coordinate among them to successfully perform the task. 
Based on the type and number of agents involved in the task the degree of difficulty of the task varies.
The decision must be made in real time and in case of an autonomous robot as the ones in the Four-Legged League, with 
limited computational resources.
Robot soccer is a particularly complex environment due to its dynamic nature resulting from the
presence of multiple teammate and opponent robots.
In particular, robots must agree on the decisions made, and who and what to do to complete the subtasks.

In general in multi-robot domains, and robot soccer in particular, collaboration 
is desired so that the group of robots work together to achieve a common
goal. It is not only important to have the agents collaborate, but also to do it
in a coordinated manner so that the task can be organized to obtain effective
results.
In this work we is address the problem of identification of collaborative behaviour in 
a Multi-Agent System (MAS) environment.
The aim is to define a systematic method to verify the effective collaboration 
among the members of a team
and compare the different multi-agent behaviours.
Using external observations of a MAS 
to analyse, model, recognize agent behaviour could be very useful to direct team actions. 
In the analysis of such systems we have deal 
with the complexity of the world state representation and with the recognition of the agent activities.
To characterise the state space, is necessary to represent temporal and spatial state changes occurred 
by agent actions.

Case-Based Reasoning (CBR) has been  successfully applied
to model the action selection of a team of robots in the robot
soccer domain~\cite{CBR_Mantaras_2009} (more precisely Four-Legged League).
A case represents a multi-robot situation where the
robots are distributed in terms of perception, reasoning, and action.
The case-based retrieval and reuse phases are based
on messages exchanged among the robots about their internal states, in terms
of beliefs and intentions.
Given a new situation, the most similar past case is retrieved and its solution
is reused after some adaptation process to match the current situation. 
The case solution is modelled as a set of sequences of actions, which indicate what
actions each robot should perform.
This case representation ensures that the solution description in the cases
indicates the actions the robots should perform; that the retrieval process
allocates robots to actions; and finally, with the coordination mechanism, that
the robots share their individual intentions to act.
This approach allows for the representation of challenging rich multi-robot actions, 
such as passes in the robot soccer domain, which require well synchronized positioning and actions.

The \textit{CBR} approach is compared with the approach presented by the Carnegie Mellon’s CMDash’06 team~\cite{Veloso05cmpack05:team}.
In their approach they have an implicit coordination
mechanism to avoid having two robots disputing the ball at the same time.
The robot in possession of the ball notifies the rest of the team, and then the
other robots move towards different directions to avoid collisions. The robots
also have roles which force them to remain within certain regions of the field
(for instance, defender, striker, etc.). The resulting behaviour of this approach
is more individualistic and reactive in the sense that the robots always try to
go after the ball as fast as possible and move alone towards the attacking goal.
Although they try to avoid opponents (turning before kicking, or dribbling),
they do not perform explicit passes between teammates to avoid them and in general they
move with the ball individually. Passes only occur by chance and are not 
previously planned. Henceforward we will refer to this approach as the \textit{reactive}
approach.

This report is addresses the problem of learning and symbolically representing the sequences of actions
performed by a teams of robots. 
The resulting relational representation of teams behaviours
enable humans to understand and study the action concepts of the observed multi-agent systems
and the underlying behavioural principles related to the complex changes of state space.
A relational sequence could be used as a qualitative representation of a team behaviour.
Low-level concepts of behaviour (events) are recognized and by these, 
high-level concepts (actions) are defined.
Our proposal is to extract from raw multi-agent observations (log files) of a
dynamic  and  complex  environment,  a  set  of  relational  sequences
describing a team  behaviour.
The method is able to discover strategic events and through the temporal 
relations between them, learns interesting actions. 
The use of relational representations in this context offers many advantages. 
One of these is generalization across objects and positions.
The set of the relational sequences has been used  to mine the most frequent pattern. 
This reduced set represents the common sequences of actions performed by the team and 
represents the characteristic behaviour of a team.
We also aim at selecting the most discriminative pattern to distinguish the reactive behaviour 
from the \textit{CBR} behaviour.

\begin{figure*}[t]
\centering
      
        {\includegraphics[width=12cm]{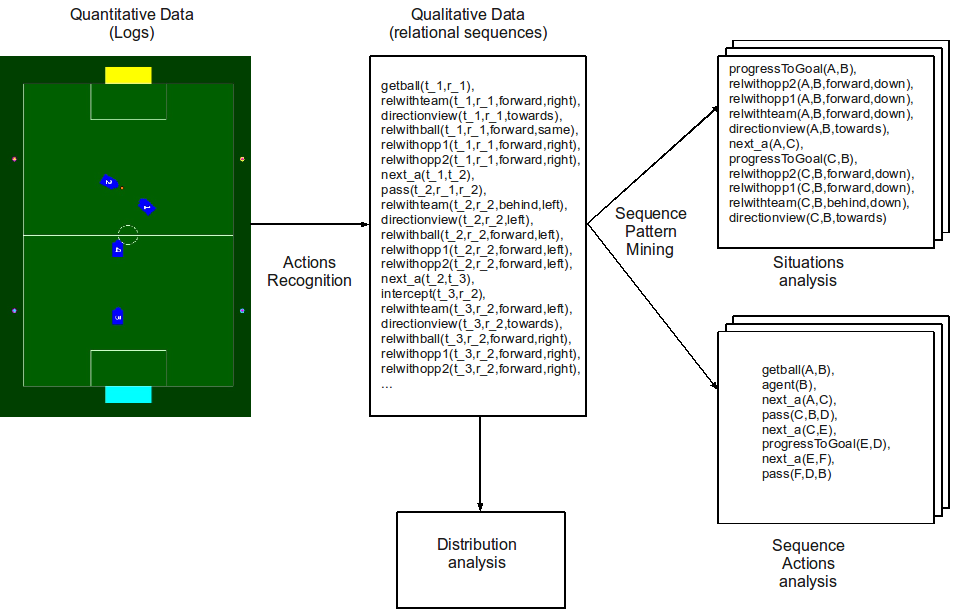}}
        
        \caption{Overview: generation of the relational sequences and the general phases of the analysis. 
                  }
        \label{fig:gen}
 \end{figure*}

\section{Case Based Reasoning approach for action selection in the Robot Soccer Domain}

Using \textit{CBR} technique, the robots are able to
perform explicit passes guided through cases.
In \textit{CBR} new problems are solved by reusing and if is necessary adapting 
the solutions to similar problems that 
were solved in the past. 
A case represents a snapshot of the environment at a given
time from a single robot point of view.
This robot is called the
\textit{reference} robot, since the information in the case is based on
its perception and internal state (its beliefs). 
The case definition is composed of three parts: the problem description,
which corresponds to the state of the game; the knowledge
description, which contains additional information used to
retrieve the case; and finally, the solution description, which
indicates the sequence of actions the robots should perform
to solve the problem. 
More formally  a case is defined as a $3$-tuple:
$$case = ((R, B, G, T m, Opp), K, A)$$
where:

\begin{enumerate}
\item $R$: relative position wrt the ball and heading of the reference robot.
\item $B$: ball’s global position.
\item $G$: defending goal.
\item $Tm$: teammates’ relative positions wrt the ball.
\item $Opp$: opponents’ relative positions wrt the ball.
\item $K$: scope of the case defined as the regions of the field
           within which the ball and the opponents should be positioned 
in order to retrieve that case. With this representation, imprecision is  easily 
handled  since positions are relate to the regions
instead of exact locations in the field.
\item $A$: sequence of actions (gameplays) each robot performs.
\end{enumerate}
 
The first step in \textit{CBR} is the retrieval of past similar cases
in order to reuse the solution of one of the retrieved cases.
Similarity is evaluated along two important measures: the
similarity between the problem and the case, and the cost of
adapting the problem to the case. Thus, the features in the 
problem description are separated into two sets: 
\textit{controllable} indices and \textit{non-controllable} indices. 
The former refers to the \textit{reference} robot and 
teammates positions (since they
can move to more appropriate positions called \textit{adapted} positions), while the latter
refers to the balls and opponents position, and the defending
goal, time and score (which is not possible directly modify). The
idea of separating the features is that a case can be retrieved
if we can modify part of the current problem description in
order to adapt it to the description of the case.

The two main functions to evaluate cases are the following:

\textit{Similarity function}: This measure indicates how similar
the non-controllable features are between the problem and
the case. Different functions for each domain of
features are defined and  then  the overall similarity is computed using the
harmonic mean of the individual similarities.

\textit{Cost function}: This measure computes the cost of modifying the 
controllable features, i.e. the cost of adapting the
problem to the case. It is defined as the sum of the distances
between the positions of the robots in the problem and the
adapted positions specified in the case after obtaining their
correspondences. The adapted positions correspond to the
global locations where the robots should be positioned in
order to execute the solution of the case.

A subset of cases are manually created and stored in a file. When the
system loads them, for each case the system automatically
generates three more cases through spatial transformations
taking into account the symmetries of the field. 
Since the considered domain is real time  and because of computational
limitations in the robots, it is essential to minimize the time
invested during the retrieval process. To speed up the search,
an indexed list is used to store the cases. Thus, given a new
problem it is easy to access the subset of cases ($CB^s$).
Cases base are indexed using the value
of the defending goal (yellow or cyan) and the number of
opponents involved in each case.

After computing the similarities and costs between the
problem and the cases in $CB^s$, a list of potential
cases is obtained. To select the retrieved case, a compromise
between the similarity degree between the problem and the
case and the cost of adapting the problem to the case is considered.
Moreover, since the domain is multi-robot  (teams
of robots), the cooperation between them is stimulated as much as possible. 
Therefore, the retrieval process orders the list of potential 
cases such that the similarity and number of players involved in the
solution of the problem are maximized, while minizing the cost.
The multi-robot system is composed of $n$ robots. All
robots interact with the environment and among themselves,
i.e. they perceive the world, they perform actions and they
send messages to each other to coordinate and to exchange
information about their internal states. Each robot has a copy
of the same case base so they can gather the information
needed during the case reuse.

Given a new state of the environment the first step is to 
select the robot responsible for the retrieval process and for the
coordination of the robots during the case reuse. 
This robot is called  the \textit{coordinator}.
The selection is based on the
distance between the robots and the ball. The further a robot
is from an object, the higher the imprecision about the 
object’s information. Therefore, the coordinator corresponds
to the one closer to the ball. Next, the \textit{coordinator} retrieves
a case according to the process described before and informs
the rest of the robots which case to reuse.

At this point the case execution begins. Firstly, all robots
that take part of the solution of the case move to their
adapted positions. Once they reach them, they send a 
message to the coordinator in order to synchronize the 
beginning of the gameplay execution with the rest of the robots.
Next, they all execute their actions until ending their 
sequences. Finally, they report the coordinator that they 
finished the execution and wait for the rest of the robots to end.
When the coordinator receives all messages, it informs the
robots so they all go back to the initial state of the process,
i.e. selecting a new coordinator, retrieving a case and 
executing its solution. The robots may abort the execution of a
case at any moment if any of the robots either detects that
the retrieved case is not applicable anymore or an expected
message does not arrive. In either case, the robot sends an
aborting message to the rest of the robots so they all stop
executing their actions and restart the process.

\section{Relational Sequential Pattern Mining}
\label{sec:RelationalSequentialPatterns}

In  this section  we  present  a method  based  on relational  pattern
mining, to extract meaningful frequent patterns able to define a behavioural 
team model.
Here we briefly review the used representation language for the domain and 
induced knowledge. For a more comprehensive introduction to logic programming
and Inductive Logic Programming (ILP) we refer the reader to ~\cite{LavracILP}.

A relational sequence is represented by a set of
Datalog~\cite{ullman88} atoms.
A first-order \emph{alphabet} consist of    a   set    of   \textit{constants},   a    set   of
\textit{variables},  a   set  of  \textit{function   symbols},  and  a
non-empty set of \textit{predicate symbols}.
 Each function symbol and 
each predicate  symbol has an \textit{arity},  representing the number
of arguments  the function/predicate has.  Constants may  be viewed as
function symbols of arity $0$.
An  atom $p(t_1,  \ldots, t_n)$  (or  atomic formula)  is a  predicate
symbol $p$ of  arity $n$ applied to $n$ terms  $t_i$ (i.e., a constant
symbol, a variable symbols, or  an $n$-ary function symbol $f$ applied
to  n  terms  $t_1,  t_2,  \ldots,  t_n$).  A  \emph{ground  term}  or
\emph{ground atom} is one that does not contain any variables. 
A  \emph{clause} is a  formula of  the form  $\forall X_1  \forall X_2
\ldots \forall X_n (L_1 \vee  L_2 \vee \ldots \vee \overline{L}_i \vee
\overline{L}_{i+1} \vee \ldots  \vee \overline{L}_m)$ where each $L_i$
is  a  literal  and $X_1,  X_2,  \ldots  X_n$  are all  the  variables
occurring  in $L_1  \vee L_2  \vee \ldots  \overline{L}_i  \vee \ldots
\overline{L}_m$.  Most  commonly the  same  clause  is  written as  an
implication $L_1, L_2, \ldots  L_{i-1} \leftarrow L_i, L_{i+1}, \ldots
L_m$, where  $L_1, L_2,  \ldots L_{i-1}$ is  the \textit{head}  of the
clause  and $L_i,  L_{i+1}, \ldots  L_m$ is  the \textit{body}  of the
clause. 
Clauses, literals and terms are said to be $\emph{ground}$ whenever they
do  not contain  variables. 
A \textit{Horn clause} is a clause which contains at most one positive literal.
A \textit{Datalog clause} is a clause with no function symbols of non-zero arity;
only variables and constants can be used as predicate arguments.

A \emph{substitution} $\theta$ is defined  as a set of bindings $\{X_1
\leftarrow a_1,  \ldots, X_n  \leftarrow a_n\}$ where  $X_i, 1  \leq i
\leq  n$ is  a variable  and  $a_i, 1  \leq i  \leq  n$ is  a term.  A
substitution $\theta$  is applicable  to an expression  $e$, obtaining
the expression $e\theta$, by  replacing all variables $X_i$ with their
corresponding terms $a_i$. 
A  conjunction  \emph{A}   is  $\theta$-$subsumed$  by  a  conjunction
\emph{B}, denoted by  \emph{$A \preceq _\theta B$}, if  there exists a
substitution \emph{$\theta$} such that \emph{$A \theta \subseteq B$}. 
A clause  $c_1$ \textit{$\theta$-subsumes} a clause $c_2$  if and only
if  there  exists  a  substitution  $\sigma$  such  that  $c_1  \sigma
\subseteq c_2$. $c_1$ is a \textit{generalization} of $c_2$ (and $c_2$
a  \textit{specialization} of  $c_1$) under  $\theta$-subsumption.  If
$c_1$ $\theta$-subsumes $c_2$ then $c_1 \models c_2$.

A \emph{relational  sequence} is  an ordered list  of atoms.   Given a
sequence $\sigma = (s_1 s_2  ...  s_m)$, a sequence $\sigma$$' = (s_1'
s_2' ...   s_k')$ is a  \emph{subsequence} (or \emph{pattern})  of the
sequence $\sigma$, indicated by $\sigma' \sqsubseteq \sigma$, if 
\begin{enumerate}
	\item $1 \leq k \leq m$;
	\item $\exists j, 1 \leq j \leq m-k$ and a substitution $\theta$
          s.t. $\forall i, 1 \leq i \leq k$: $s_i' \theta = s_{j+i}$.
\end{enumerate}
A subsequence  occur in a sequence  if exists at least  a mapping from
elements  of $\sigma$'  into the  element  of $\sigma$  such that  the
previous  condition are  hold.  In  our case,  that  subsequence is  a
\textsl{relational pattern}.

The \emph{support}  of a
sequence $\sigma$  in a set of sequences  $\mathcal{S}$ corresponds to
the  number  of sequences  in  $\mathcal{S}$  containing the  sequence
$\sigma$:  support($\sigma$)= $|\{ \sigma'  | \sigma'  \in \mathcal{S}
\wedge \sigma \sqsubseteq \sigma' \}|$.

If we consider a sequence as an ordered succession of events, in our description language
we distinguish two kinds of Datalog atoms: \textit{dimensional} and \textit{non-dimensional} atoms.
A dimensional atom explicitly refers to dimensional relations between events involved in the sequence. 
A non-dimensional atom denotes relations between objects (with arity greater than 1), or 
characterizes an object (with arity 1) involved in the sequence.
In order to represent relational patterns, a dimensional operator must be introduced.
We use  \texttt{next/2} to denote the direct successor operator.
For instance, \texttt{next(x,y)} denotes that the event y is the direct successor of the event x.

In order to mining the most frequent patterns, we use an Inductive Logic
Programming (ILP)~\cite{muggleton94}   algorithm,    based   on
\cite{esposito08fundamenta},
 for  discovering relational patterns from
sequences. It  is based on a  level-wise search method,  known in data
mining  from the  \textsc{Apriori}  algorithm~\cite{AgrawalEtAl96}. It
takes into account the sequences, tagged with the belonging class, and
the $\alpha$ parameter denoting the minimum support of the patterns.
It is  essentially composed by  two steps, one for  generating pattern
candidates and the other for evaluating their support. 
 The level-wise  algorithm makes a breadth-first search  in the lattice
of patterns ordered by a specialization relation. 
Starting from the most general  patterns, at each level of the lattice
the algorithm generates candidates  by using the lattice structure and
then evaluates the frequencies of the candidates.  
In this phase some patterns may be discarded due to the monotonicity  of pattern  frequency 
(if a  pattern is  not frequent then none of its  specializations is frequent).

The generation  of the patterns  actually present in the  sequences of
the dataset is based on a top-down approach.  
The  algorithm starts with  the most  general patterns.  These initial
patterns are  all of length 1 and  are generated by adding  an atom to
the empty pattern. Then, at each step it tries to
specialize all  the potential  patterns, discarding those that
do not occur in any sequence and storing the ones whose length is equal
to the user specified input parameter \emph{maxsize}.
Furthermore,  for each  new refined  pattern,  semantically equivalent
patterns are  detected by using  the $\theta$-subsumption relation
and  discarded.  In   the  specialisation  phase,  the  specialisation
operator under $\theta$-subsumption is used. Basically, the operator
adds atoms to the pattern. 
Finally, the  algorithm may use a background  knowledge $\mathcal{B}$ (a
set of Datalog  clauses) containing constraints on how  to explore the
lattice.

\section{Learning  behavioural relational  representation}

This section provides  a description of the approach that we use to extract relational sequences 
from log files, which are able to describe and characterise the behaviour of a team of agents.

\begin{figure*}[t]

\centering

        {\includegraphics[width=6cm]{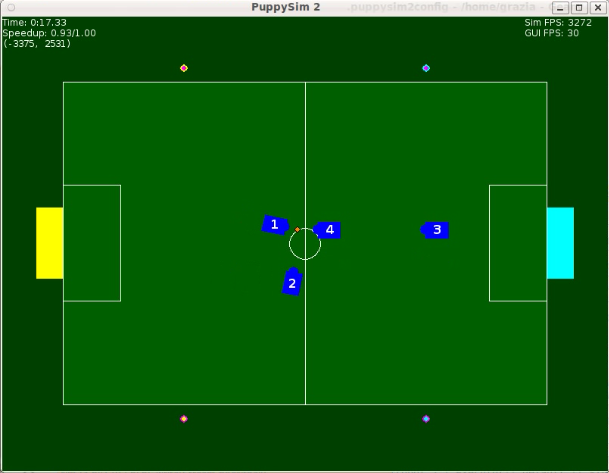}}
        {\includegraphics[width=6cm]{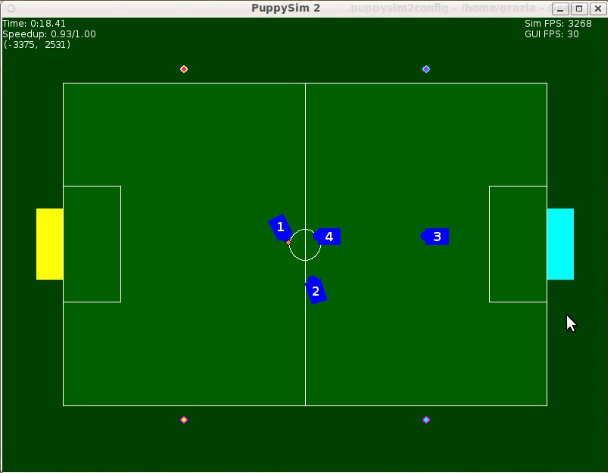}}

        {\includegraphics[width=6cm]{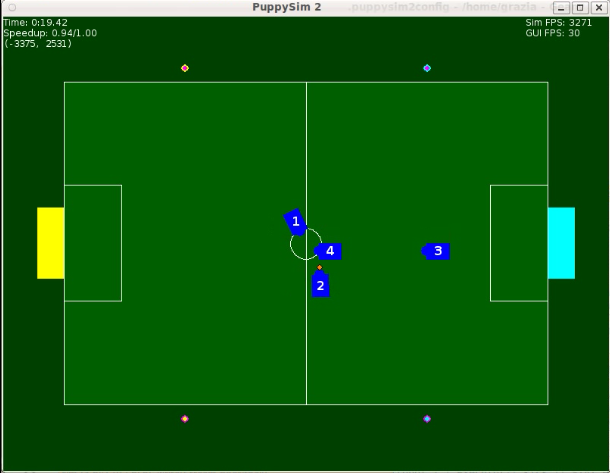}}
        {\includegraphics[width=6cm]{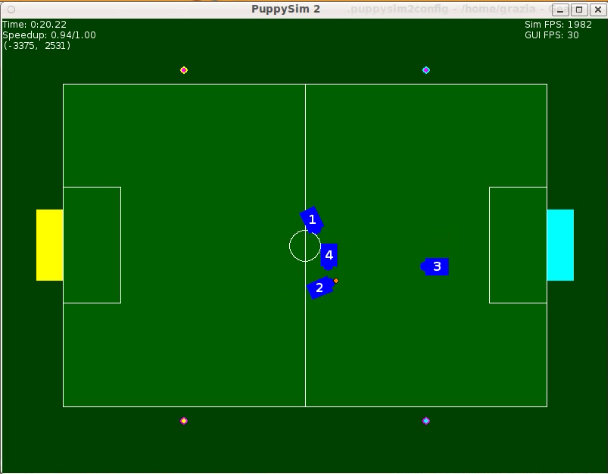}}
        \caption{An example of a sequence of actions}
        \label{fig:seqaction}
 \end{figure*}

An example of a sequence of actions is presented in
Figure~\ref{fig:seqaction}. The four frames show a successful sequence 
of actions to  overcome an opponent.
The robot players are represented as blue boxes with a semicircular shape
indicating body orientation. The attacking team is composed of robots 1 and 2, and the
defending team of robots 3 and 4 (where robot 3 is the goalkeeper). 
The ball correspond to the orange circle.

The log used represents a stream of consecutive \textit{raw observations} about each soccer 
player's position and 
the position of the ball at each moment of the trial.
From this log streams it is possible to  recognize basic actions (\textit{high-level concepts}).
Each  team has sequences of basic actions used  to form  coordinated  activities which  attempt 
to achieve  the team's goals.  
In our work, we identify the following basic actions of the players:\\
\begin{itemize}
 \item  $\textbf{getball}(T,Player_n)$:  
  at time $T$, $Player_n$ gains possession of the ball;\\
\item $\textbf{catch}(T,Player_n)$:  
  at time $T$, $Player_n$ gains possession of the ball previously belonging to an opponent;\\
\item $\textbf{pass}(T,Player_n,Player_m)$:  
  $Player_n$ kicks the ball and at time $T$ the $Player_m$ gains possession, and both players are from the same team;\\
\item $\textbf{dribbling}(T,Player_n)$:  
  at time $T$, $Player_n$ moves a significant distance avoiding an opponent;\\
\item $\textbf{progressToGoal}(T,Player_n)$:  
  at time $T$, $Player_n$ moves with the ball toward the the penalty box;\\
\item $\textbf{aloneProgressToGoal}(T,Player_n)$: 
  at time $T$, $Player_n$ moves alone with the ball toward the penalty box, without teammate between it and the goal area;\\
\item $\textbf{intercept}(T,Player_n)$: 
  at time $T$, $Player_n$ loses the possession of the ball, and the new owner of the ball is from the opponent team;\\
\end{itemize}

The log stream is processed to infer the low-level $events$ that occurred during a trial.
An event takes place when the ball possession changes or the ball is out of bounds. 
Each next recognized event performed by a team contributes to form an action.
To better  describe the behaviour of an entire team, it is necessary to take into 
account the state of the word and the time in which the action is performed.
Agents in dynamic environments have to deal with world representations that changes over time.
A qualitative description of the world allows a concise and powerful representation of the relevant
information.

Each recognized event has some persistence over time and remains active 
until another event incompatible with it occurs.
For example when a robot takes ball possession, an event indicating a new possession is generated.
Subsequently, if an opponent manages to take possession of the ball, this generates a new event not 
compatible with the previous one and therefore it expires.
An event that occurs in parallel with another event is called a \textit{contemporary} event.
For example, if an opponent tries to steal the ball from a robot that is in possession of it,
the event of ``trying to steal the ball'' is contemporary to the event ``ball possession''.
This \textit{contemporary} event holds until one of the robots is able to take full possession of the ball, 
(i.e. moves away with the ball) or when the ball goes out of bounds.

A set of recognized events contributes to define an action.
For example, when a player catches the ball, it could be due to a pass,  an interception, or a 
dribbling, depending on the previous events still active.
When an opponent tries to catch the ball, then  a $contemporary$ event occurs.
In this case, if the ball did not belong to any robot for a while, then the system considers this 
situation to be a ``$getball$''.
If the previous event was ``catch of the ball'' by the same robot and no opponent
has attempted to take possession, then the system considers it a ``$aloneProgressToGoal$'', or a
``$progressToGoal$'' or a ``$getball$'' event, depending on the state of the world.
In this situation, if a $contemporary$ event is still held the system recognizes a ``$dribbling$''. 
If the previous event was the gain of the ball possession by an opponent, 
the the system considers this a ``$catch$''.
 
The current world state is represented by the positions of the players (teammates and opponents),
and the  ball.
Instead of describing these through the actual coordinates or the identification of an area of the field,
is more useful to describe the relation between these components of the state world.
In this context, to characterise adequately specific scenes, we considered
the viewpoint of the robot that performs the  action, to determine how it interacts with others.
This egocentric viewpoint has to rely on simple distinctions as, for instance, the position 
of another robot (a teammate or an opponent) that could be left or right of it, in front or 
back of it. 
More precise and objective descriptions would not reflect the generality and 
would not allow to abstract to positional information.
Sequences represent  a symbolic abstraction of the  raw observation.

In particular, to describe the relation $\textit{direction\_view}$ of the player 
with respect to the opponent's penalty box, 
we use $front$, $left$, $right$, $backwards$.

To describe the relation of the player with respect to the teammate, the ball and the opponents,
we have used two arguments, one for the ``horizontal'' relation ($forward$ or $behind$) 
and the other for the ``vertical'' relation ($left$ or $right$).
We use $same$ when the player has the same position with respect to the teammate, the ball and the opponents.

\begin{itemize}
\item $\textbf{direction\_view}(T,Player_n,position)$;\\
\item $\textbf{rel\_with\_ball}(T,Player_n,horizontal,vertical)$;\\
\item $\textbf{rel\_with\_team}(T,Player_n,horizontal,vertical)$;\\ 
\item $\textbf{rel\_with\_opp1}(T,Player_n,horizontal,vertical)$;\\
\item $\textbf{rel\_with\_opp2}(T,Player_n,horizontal,vertical)$;\\
\end{itemize}

\begin{figure}[t]
        \centering
        {\includegraphics[width=7cm]{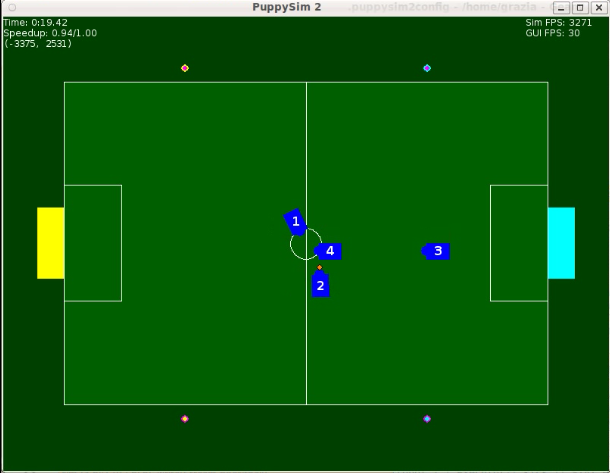}}        
        \caption{ }
        \label{fig2:exworld}
 \end{figure}

For instance, the following predicate describes the environment depicted the Figure~\ref{fig2:exworld}:\\
\noindent 
pass($time_n,robot_1,robot_2$),\\
direction\_view($time_n,robot_2,right$),\\
rel\_with\_team($time_n,robot_2,forward,left$),\\
rel\_with\_ball($time_n,robot_2,forward,same$),\\
rel\_with\_opp1($time_n,robot_2,forward,right$),\\
rel\_with\_opp2($time_n,robot_2,forward,right$)\\
\noindent
where opp1 represents robot 3, and opp2 represents robot 4.

The following predicates describe the result of the trial:
\begin{itemize}
\item $\textbf{goal}(T)$ : at time $T$ the ball enters in the opponent's goal.\\ 
\item $\textbf{to\_goal}(T)$ : at time $T$ goes out of the field but passes near one of the goal posts.\\
\item $\textbf{ball\_out}(T)$ : at time $T$ the ball goes out of the field without being a goal or close to goal.\\
\item $\textbf{block}(T)$ :  at time $T$ the goalie stops or kicks the ball.\\
\item $\textbf{out\_of\_time}(T)$ : time out.\\
\end{itemize}

This is an example of a sequence of actions and the environment description:\\

\noindent
getball($time_1,robot_1$),

rel\_with\_team($time_1,robot_1,forward,right$),\\
 \indent direction\_view($time_1,robot_1,front$), \\
 \indent rel\_with\_ball($time_1,robot_1,forward,right$),\\ 
 \indent rel\_with\_opp1($time_1,robot_1,forward,right$),\\ 
 \indent rel\_with\_opp2($time_1,robot_1,forward,right$), \\
 \indent next\_a($time_1,time_2$), 
 
\noindent
pass($time_2,robot_1,robot_2$), 
 
rel\_with\_team($time_2,robot_2,behind,left$), \\
 \indent direction\_view($time_2,robot_2,left$), \\
 \indent rel\_with\_ball($time_2,robot_2,forward,left$),\\ 
 \indent rel\_with\_opp1($time_2,robot_2,forward,left$), \\
 \indent rel\_with\_opp2($time_2,robot_2,forward,left$), \\
 \indent next\_a($time_2,time_3$), 

\noindent
intercept($time_3,robot_2$), 

rel\_with\_team($time_3,robot_2,forward,left$), \\
 \indent direction\_view($time_3,robot_2,front$), \\
 \indent rel\_with\_ball($time_3,robot_2,forward,right$), \\
 \indent rel\_with\_opp1($time_3,robot_2,forward,right$), \\
 \indent rel\_with\_opp2($time_3,robot_2,forward,right$), 

\noindent
agent($robot_1$), \\
agent($robot_2$), \\
opponent($op_1$), \\
opponent($op_2$)\\

\noindent
where $next_a(time_{n_1},time_{n_2})$ define the temporal relation between the actions.

\section{Experimental evaluation}
\normalsize

The aim of this experimentation is to measure and demonstrate the degree of collaboration 
of teams of soccer playing robots and, from a more general point of view, to characterise a Multi-Agent System behaviour.
With the \textit{CBR} approach the performance of the robots should result in a cooperative behaviour
where the team works together to achieve a common goal, a desired property in this
kind of domain. 
Using a relational representation on the actions actually performed by a team,
we intend to evaluate the collaborative behaviour of the team.
Through the pattern mining method used, the most frequent set of behaviours is extracted.
This set is able to characterise the behaviour of an entire team.

The  \textit{CBR} approach allows the robots to apply a more deliberative
strategy, where they can reason about the state of the game in a more global
way, as well as to take into account the opponents. Thus, they try to avoid
the opponents by passing the ball to teammates, which should increase the
possession of the ball, and therefore, the team should have more chances to
reach the attacking goal.
As we will see, experiments revealed that these action sequences characterise the behaviour of the \textit{CBR} team.
To be more precise, these action patterns are in the set of most significant patterns extracted 
from \textit{CBR} team sequences, whereas they are not among the most significant patterns extracted 
from the reactive (\textit{REA}) team sequences.

We used an extended version of the PuppySim 2 simulator~\cite{CBR_Mantaras_2009}, 
created by the CMDash team~\cite{Veloso05cmpack05:team}. 
This simulator represents the basic aspects of the RoboCup Standard Platform League, 
Four-Legged Soccer Competition~\cite{fourleggedleaguerulebook}.
Some additional features were implemented, such as managing team messages, robots walking
while grabbing the ball, etc. The final version of the simulator is a simplified
version of the real world. The robots’ perception is noiseless, i.e. the ball’s
position and the location of all robots on the field is accurate. However the
outcome of the actions the robots perform have a certain degree of randomness.
The kicks are not perfect and the ball can end up in different locations within
or around its ideal trajectory. In addition, when the robot tries to
get the ball, it does not always succeed, simulating a ``grabbing'' failure (a
very common situation with the real robots). The ball’s movement is modelled
taking into account friction, starting with a high speed and
gradually decreasing until stopping (if no one intercepts it before).

The case base used into the experimentation is composed of 136 cases.
From this set, 34 cases are hand-coded,
while the remaining ones are automatically generated using spatial transformations 
exploiting the symmetries of the soccer field.

The robots using
the \textit{CBR} approach perform a default behaviour when no case is found. In these
experiments, the default behaviour corresponds to the reactive approach.
A simple behaviour for the opponents (defender, midfield
defender and goalie) was implemented. Each robot has a home region and it cannot go beyond
that region. If the ball is within its home region, then the robot moves 
towards the ball and clears it. Otherwise, the robot remains in the boundary
of its home region, facing the ball to maintain it in its field of view. The
experiments consist of two vs. two games.

\subsection{Setting}

Two sets of simulated experiments, one with the \textit{CBR} team and another one with the
\textit{REA} team, were performed taking into account two different scenarios.
Besides, two possible configurations for the opponents are defined.
The first is called DG configuration and considers a defender and a goalie.
The second one, the 2D configuration, correspond to a midfield defender and a defender. 
The penalty area is reserved for the goalie, thus defenders are not allowed to enter into it.

In the 2D configuration each defender has its own home region with an overlapping area.
The strategy of assigning regions to players is commonly used in robot soccer teams.
In this way all the regions on the field are covered by at least one player and 
and the situation in which all robots chasing the ball is avoided.

\begin{figure*}[t]
\centering
       
        \subfigure[]{\includegraphics[width=6cm]{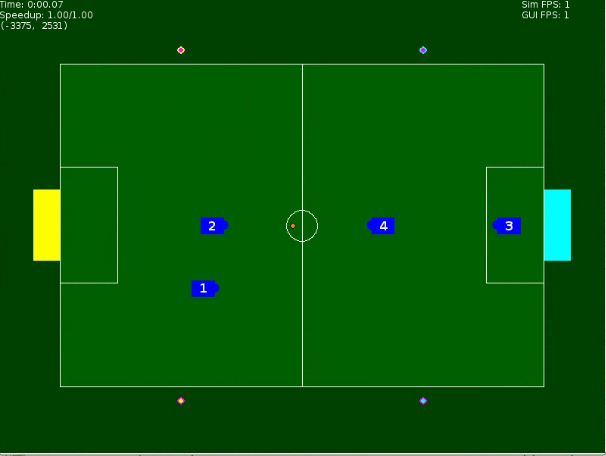}}
        \subfigure[]{\includegraphics[width=6cm]{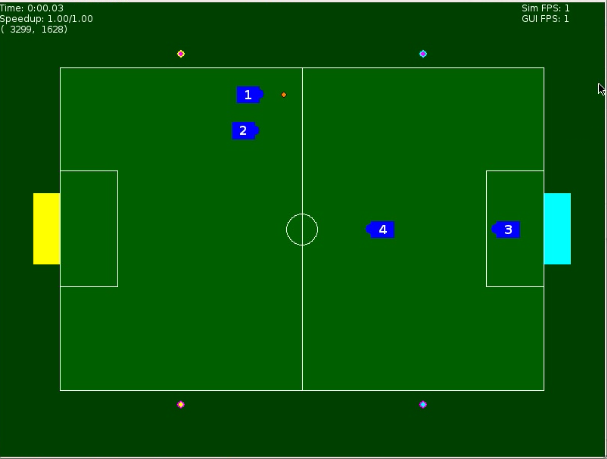}}
        \hspace{5mm}
        \subfigure[]{\includegraphics[width=6cm]{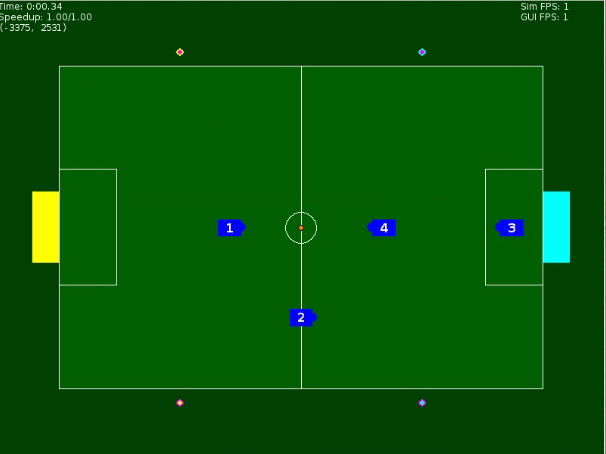}}
        \subfigure[]{\includegraphics[width=6cm]{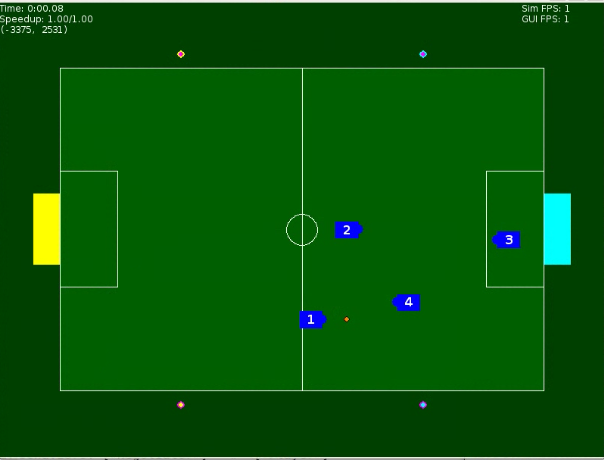}}
        \caption{Scenarios used during the experimentations with the DG configurations.}
        \label{fig:scenarios}
 \end{figure*}

Four basic scenarios have been defined.
These scenarios correspond to typical situations that usually occurs in football matches,
where the attackers are coming from the back of the field towards the
attacking goal, while the opponents are waiting at their positions.
Each scenario is used with both configurations of opponents (DG or 2D). 
In scenario 1 (Figure~\ref{fig:scenarios}a)
the ball and the attackers (robots 1 and 2) are positioned in the 
middle-back of the field, while in scenario 2 (Figure~\ref{fig:scenarios}b), 
they are located in the left side of the field. 
The opponents (goalie, 3, and defender, 4, in these figures)
remain within their home region.
In the 2D configuration, the goalie is replaced with a defender.
 
In scenarios 3 and 4 (Figure~\ref{fig:scenarios}c and Figure~\ref{fig:scenarios}d), 
the ball and attackers are located in the middle-front of the field. 
From a strategic point of view these scenarios are more interesting.
The first decision (action) the attackers make
(execute) is critical in their aim to reach the goal while avoiding the defender(s)
whose main task is to intercept or to steal the ball.

These two sets of scenarios are general enough to represent the most 
important and qualitatively different situations the robots can encounter in a game.
Initiating the trials on the left or right side of the field does not make much 
difference with respect to the actions the robots might perform in any of the two evaluated
approaches, since they would perform the corresponding symmetric actions
instead. 
We are interested in the defenders being active opponents complicating the attackers’ task.
Thus, each scenario has been defined with the ball near the attacking goal,
to allow defenders to attack and steal the ball.

The starting setting is fixed for any experimentation trial, but due 
imprecision of the actions the robots perform, 
the development of the trials varies from one to another.
For this reason, during a trial very different and unpredictable 
situations may occur not known in advance.

\subsection{Results}
\label{sec:experiment}

In order to evaluate our approach we analyse log files of soccer games.
We have implemented a system that is able to identify and extract the interesting high-level concepts and 
construct sequences of coordinated team behaviours using the recorded observations (logs) of this simulation games. 
The sequences have been defined on $7$ atomic behaviours (catch, pass, dribble, etc.) 
and $5$ environment descriptions (rel\_with\_ball, rel\_with\_team, etc.).

We performed $500$ trials for each approach (\textit{CBR} and \textit{REA}) and each scenario in the DG configuration, 
for a total of $4000$ trials.
From the raw observations of the log files we have obtained the dataset corresponding to this configurations.
It is made up of $10261$ sequences ($6242$ sequences \textit{CBR} approach and $4019$ 
sequence for the reactive behaviour \textit{REA}). 

Regarding the 2D configuration, we observed that the time required to end a trial was too long.
This was due to the ability of the two defenders in preventing the attackers to reach the goal, not allowing them to reach the goal.
For this reason a timeout to 60 second to end the trial was adopted.
For the 2D configuration, we have performed $200$ trials per scenario and per approach, 
for a total of $1600$ trials.
The dataset is made up of $4329$ sequences($2392$ sequences \textit{CBR} approach and $1937$ 
sequence for the reactive behaviour \textit{REA}).

Data has analysed considering three dimensions:
one that takes into account the distribution of actions recognized,
a second one considering only the predicates that represent actions, 
and the last dimension, related to the environment in which actions were performed.

\subsection{Distribution of the actions} 

\begin{figure*}[t]
\centering
       
        \subfigure[DG configuration]{\includegraphics[width=6cm]{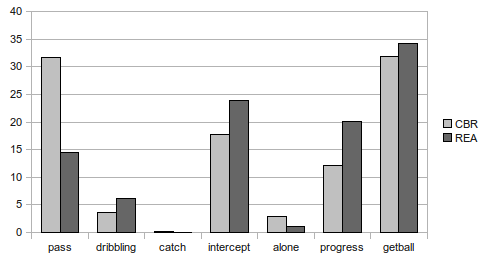}}
        \subfigure[2D configuration]{\includegraphics[width=6cm]{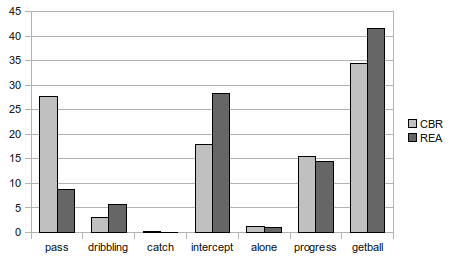}}
        
        \caption{Distribution in percentage of the actions on the entire dataset per configurations}
        \label{fig:distr}
 \end{figure*}

In this section we analyse the distribution of the action recognized from the logs to better understand
the behaviour of the entire team. 
Simply observing the composition of sequences is possible to make preliminary observations related
to the degree of collaboration among robots.
Modelling  multi agent behaviour through relational sequences allows a good level of abstraction.
Indeed, the distribution, in percentage, of actions recognized in all scenarios
has the same characteristics (Figure~\ref{fig:distr}) taking into account only the recognized actions.
In particular, this indicates that with this approach it is possible to define a model of the actual team behaviour.

As we can see in Table~\ref{table:DG} and Table~\ref{table:2D}  the amount of sequences able to describe the
behaviour of the team that uses the \textit{CBR} approach is significantly higher than the one that uses 
the reactive approach.
Since the \textit{CBR} team plays using collaborative strategies, where the robots try to reach the goal area
usually by passing the ball to a teammate, or moving to adapted positions to reuse
the selected case, more sequences and therefore more actions are needed to describe such behaviour.

When the robot holding the ball tries to move towards the penalty area, 
having in front an opponent, it can act in a cooperative or individualistic way.
That is, it can pass the ball to its teammate (in this case the recognized actions would be $getball$ 
and $pass$) or could simply try to overpass the opponent, 
adopting an individualistic behaviour (if the robot succeeds in its aim,
the action recognized will be a $dribbling$).

\begin{table}[t]
\caption{Recognized high-level concepts on DG configuration.} 
\label{table:DG} 
\centering
\begin{tabular}{|c|cc|cc|cc|cc|} 
\hline
& 
\multicolumn{2}{|c|}{scenario A}&
\multicolumn{2}{c|}{scenario B}&
\multicolumn{2}{c|}{scenario C}&
\multicolumn{2}{c|}{scenario D}\\
   & CBR & REA & CBR & REA & CBR & REA & CBR & REA \\
\hline 

 N. sequences & 1595 & 977 & 1513 & 1170 & 1623 & 837 & 1511 & 1035\\
\hline
pass &2285 & 1325 & 3135 & 557 & 2293 & 47 & 2023 & 190\\
dribbling &256 & 217 & 254 & 234 & 242 & 161 & 334 & 282\\
catch &5 & 3 & 24 & 3 & 10 & 0 & 7 & 0\\
intercept &1385 & 857 & 1306 & 1003 & 1434 & 771 & 1324 & 840\\
aloneProgressToGoal &216 & 35 & 291 & 44 & 192 & 34 & 177 & 34\\
progressToGoal &1261 & 585 & 974 & 1200 & 981 & 662 & 511 & 481\\
getball &2583 & 1246 & 2467 & 1423 & 2461 & 947 & 2272 & 1358\\
\hline
  tot. Actions &7991 & 4268 & 8451 & 4464 & 7613 & 2622 & 6648 & 3185\\

\hline
\end{tabular}
\end{table}

\begin{table}[t]
\caption{Recognized high-level concepts on 2D configuration.} 
\label{table:2D} 
\centering
\begin{tabular}{|c|cc|cc|cc|cc|} 
\hline
& 
\multicolumn{2}{|c|}{scenario A}&
\multicolumn{2}{c|}{scenario B}&
\multicolumn{2}{c|}{scenario C}&
\multicolumn{2}{c|}{scenario D}\\
   & CBR & REA & CBR & REA & CBR & REA & CBR & REA \\
\hline  
N. sequences & 622 & 449 & 598 & 477 & 570 & 543 & 602 & 468\\
\hline 
pass & 570 & 107 & 865 & 199 & 613 & 95 & 769 & 92\\
dribbling & 77 & 83 & 85 & 73 & 84 & 65 & 63 & 99\\
catch & 4 & 0 & 6 & 0 & 4 & 0 & 3 & 2\\
intercept & 468 & 373 & 460 & 389 & 424 & 468 & 467 & 346\\
alone & 34 & 9 & 24 & 17 & 34 & 13 & 40 & 22\\
aloneProgressToGoal & 342 & 191 & 410 & 350 & 352 & 225 & 459 & 38\\
progressToGoal & 883 & 508 & 907 & 605 & 801 & 682 & 896 & 524\\
\hline 
tot. Actions & 2378 & 1271 & 2757 & 1633 & 2312 & 1548 & 2697 & 1123\\

\hline
\end{tabular}
\end{table}

As can clearly seen in Figure~\ref{fig:distr}  the number of the $pass$ actions  that appear in  
\textit{CBR} sequences is significantly higher than in \textit{REA}  sequences.
The number of the $dribbling$ actions  that appear in the \textit{REA} sequences is higher than in  
\textit{CBR}  sequences.
Actions like  $getball$, $progressToGoal$ and $aloneProgressToGoal$ indicate progress towards 
the goal area. The first two are more frequent in the \textit{REA} team and this denotes  a 
individualistic behaviour.
Instead, actions such $aloneProgressToGoal$ are more frequent in the case of \textit{CBR} sequences 
because, thanks to collaborative behaviour by means of passing, robots can overcome the defenders 
and progress alone towards the goal area, this is the only case in which a CBR robot moves alone 
since it has no teammate with whom collaborate.
This is also true the case of 2D configuration in which we have two defenders, where the number 
of $progressToGoal$ actions recognized in \textit{CBR} sequences is larger.

\subsection{Sequence Actions Analysis} 

Here we consider the sequence analysis taking into account only the actions performed 
during the trials, without considering the predicates describing the state of the world.
The goal of this experimentation was to find a subgroup of most meaningful patterns of actions 
able to characterise the behaviour of a team.
We have used the whole dataset, all the sequences of the all scenarios per configuration.
In particular, for the $DG$ configuration we have $10261$ sequences, $6242$  of which correspond to the \textit{CBR}
and $4019$, to the \textit{REA}.
Regarding the 2D configuration, the dataset is made up of $4329$ sequences. 
In particular, we have $2392$ sequences for \textit{CBR} and  $1937$ sequence for \textit{REA}.

To select the most meaningful patterns, i.e. a subset of frequent patterns that is able to characterise
the essential behaviour of a team, we have used  as measure the Fisher Score.  
The Fischer score~\cite{Duda00} is popularly used in classification system to measure the 
discriminative power of a feature.
In general, it is defined as 
\begin{equation}\label{eq:FS}
Fr =  \frac
{\sum_{i=1}^c n_i(\mu_i - \mu)^2}
{\sum_{i=1}^c n_i \alpha_{i}^{2}}
\end{equation}
where $n_i$ is the number of data samples in class $i$, in our case the number of the sequences in class $i$, 
$\mu_i$ is the average feature value in class $i$, in our case is the average number of the occurrences of 
the pattern mined in class $i$,  $\alpha_{i}$ is the standard deviation of the feature in class $i$, and 
$\mu$ is the average feature value in the whole dataset.

In~\cite{Cheng07discriminativefrequent} Cheng et al. demonstrate the
frequency upper bound of discriminative measures such as
information gain and Fisher score, showing a relationship
between frequency and discriminative measures, and between the two discriminative measures.
Since pattern of low support have a limited coverage of the dataset, 
these have a very limited discriminative power. 
But on the other hand, patterns of very high support have also a very limited discriminative power, 
since they are too common in the data.
Therefore, in general it is appropriate to find patterns not too frequent  
with suitable support threshold.
But this implies a greater effort during the pattern mining step.

Frequent patterns reflect strong association between objects, in this case represents common
behaviours adopted by a team.
The frequency is calculated over the whole dataset and over both sets of sequences (\textit{CBR} and \textit{REA}).
Among the different sequences for both teams, the most frequent patterns belong to the
\textit{CBR} team. 
For this reason we have used the threshold  $\sigma = 0.10$, which is high enough to ensure adequate 
coverage of the dataset and sufficiently low to allow to discover frequent sequences also for the 
 \textit{REA} team.

\begin{table}[t]
\caption{Some examples of most interesting patterns mined.} 
\label{table:pattern} 
\centering
\begin{tabular}{|lcc|} 
\hline

pattern  & Fisher score  & team  \\
\hline  
getball(A,B),next\_a(A,C),pass(C,B,D)  &  0.23494427 & cbr\\
pass(A,B,C),next\_a(A,D),getball(D,C) & 0.07889064 & cbr\\
progressToGoal(A,B),next\_a(A,C),intercept(C,B) & 0.07138557 & rea\\
progressToGoal(A,B),next\_a(A,C),dribbling(C,B) &0.03037611 & rea\\
\hline 
getball(A,B),next\_a(A,C),pass(C,B,D),& & \\
next\_a(C,E), intercept(E,D) & 0.05987475 & cbr\\
getball(A,B),next\_a(A,C),getball(C,B),& & \\
next\_a(C,D),pass(D,B,E) & 0.05517990 & cbr\\
getball(A,B),,next\_a(A,C),pass(C,B,D),& & \\
next\_a(C,E),progressToGoal(E,D) & 0.04290462 & cbr\\
progressToGoal(A,B),next\_a(A,C),progressToGoal(C,B),& & \\
next\_a(C,D),intercept(D,B) & 0.03860653 & rea\\
progressToGoal(A,B),next\_a(A,C),progressToGoal(C,B),& & \\
next\_a(C,D),progressToGoal(D,B) & 0.01199806 & rea\\
\hline
getball(A,B),next\_a(A,C),pass(C,B,D),next\_a(C,E),& & \\
 progressToGoal(E,D),next\_a(E,F),pass(F,D,B) & 0.02503889 & cbr\\
getball(A,B),next\_a(A,C),pass(C,B,D),next\_a(C,E), & & \\
getball(E,D),next\_a(E,F),pass(F,D,B) & 0.01994761 & cbr\\
getball(A,B),next\_a(A,C),getball(C,B),next\_a(C,D),& & \\
 pass(D,B,E),next\_a(D,F),intercept(F,E) &  0.01829672 & cbr\\

\hline
\end{tabular}
\end{table}

Table~\ref{table:pattern} shows the most interesting patterns that have been extracted.
As we can easily see, the presence of the predicate $pass$ is enough to distinguish the \textit{CBR} team.
Indeed, this type of action indicates collaborative behaviour, and is typical of sequences that 
can characterise the team \textit{CBR}.

\subsection{Situations Analysis} 

The purpose of this analysis is to assess the state of the world when the action is performed.
At the same state of the world, with the necessary abstractions related to the positions of the teammate, 
of the opponents and of the ball  which are possible through a relational 
representation, follows a different action depending on the behaviour of the team.
Here we consider the patterns mined taking into account predicates related to the description 
of the environment in which the agent acts. A pattern mined consists of sets of predicates such as 
\textit{rel\_with\_opp2(T, Player, horizontal, vertical)}, \textit{rel\_with\_opp1(T, Player, vertical, horizontal)}, 
\textit{rel\_with\_team(T, Player, horizontal, vertical)} and so on,  at the time of the action
was performed. The mined set represents considerable information about the state of the world.
Analysing this kind of situations is possible to abstract the difference between the behaviour of the two 
teams, the \textit{CBR} and the \textit{REA}.

For example in a situation in which the robot is in front of the penalty area and the teammate 
is in front, as well as the two opponents and the next action corresponds to a pass to the teammate
 is a typical sequence for the \textit{CBR} team
with a significant Fisher Score:
 
\begin{example}
 
$getball(A, B),\\
rel\_with\_opp2(A, B, forward, down),\\
rel\_with\_opp1(A, B, forward, down),\\
direction\_view(A, B, front), \\    
next\_a(A, C), pass(C, B, D),\\
rel\_with\_opp2(C, D, forward, down),\\
rel\_with\_opp1(C, D, forward, down)$
\end{example}

In the same context, the characteristic sequence for the \textit{REA} team that would be
an  interception. That is:
\begin{example}

$getball(A, B),\\
rel\_with\_opp2(A, B, forward, up),\\  
rel\_with\_opp1(A, B, forward, up),\\
next\_a(A, C),\\
intercept(C, B),\\ 
rel\_with\_opp2(C, B, forward, up),\\
rel\_with\_opp1(C, B, forward, up),\\
direction\_view(C, B, front)$ 
\end{example}

This means that for the same situation, the \textit{CBR} team tries to overcome an opponent 
through a pass while, a team with a purely reactive behaviour tries to move towards the penalty area 
incurring more frequently in interception by the opponents.
This indicates an individualistic behaviour. 
This analysis characterises the two different team behaviours and confirms the conclusions 
of previous analysis.

\section{Conclusions}
\label{sec:conclusion}

In this report we have shown the potential of the use of  a relational representation
in a Multi-Agent domain to model the behaviour of the whole system.
In this way it is possible to define a more high-level view of the behaviour on the multi-robot systems 
using a multi-agent activity logs. 

In particular, this report is focused on the challenge of autonomous unsupervised
learning of team behaviours based on observations.
The aim was also to try to measure and demonstrate the degree of collaboration, analysing 
the joint behaviour of the teams.
A desired property in a Multi-Agent systems is the ability of the team to work together
to achieve a common goal in a cooperative manner.
Our approach uses symbolic sequences (a relational representation) identification to translate 
raw multi-agent, multi-variate observations of a dynamic, complex environment, into a set of 
sequential actions that are characteristic of the team in question.
The implemented method is able to discover strategic events and through the temporal 
relations between them learns the interesting actions.
Raw multi-agent data logs were transformed into a set of sequential symbolic actions 
able to describe the  team behaviour.

We compared the performance of two teams (REA and CBR) in the RoboCup four-legged league simulated environment, 
which have a very different behavioural approach.
In general the results obtained in experiments confirm that the recognized action sequences 
characterise the behaviour of the teams.
The set of relational sequences has been used to mine the most frequent patterns.
This reduced set represents the common sequences of actions preformed by the teams.

\appendix
\section{Appendix A}
\label{sec:AppA}

\textbf{Some examples of the most interesting patterns mined belonging to the \textit{REA} team.}
\begin{itemize}

\item progressToGoal(A, B),  rel\_with\_opp2(A, B, forward, down), rel\_with\_opp1(A, B, forward, up), rel\_with\_team(A, B, forward, up), direction\_view(A, B, front),     
next\_a(A, C), rel\_with\_opp1(C, B, forward, up), rel\_with\_team(C, B, forward, up), direction\_view(C, B, front)    0.0406468    \\
\item progressToGoal(A, B), rel\_with\_opp2(A, B, forward, down), rel\_with\_opp1(A, B, forward, up), rel\_with\_team(A, B, forward, up), direction\_view(A, B, front),     
next\_a(A, C), pass(C, B, D), rel\_with\_opp2(C, D, forward, up), direction\_view(C, D, left)    0.0404972   \\
\item progressToGoal(A, B), rel\_with\_opp2(A, B, forward, down), rel\_with\_team(A, B, forward, up), direction\_view(A, B, front),     
next\_a(A, C), pass(C, B, D), rel\_with\_opp2(C, D, forward, up), rel\_with\_opp1(C, D, forward, up), direction\_view(C, D, left)   0.0369131    \\
\item progressToGoal(A, B), rel\_with\_opp2(A, B, forward, down), rel\_with\_opp1(A, B, forward, up), rel\_with\_team(A, B, forward, up), direction\_view(A, B, front),     
next\_a(A, C), rel\_with\_team(C, B, forward, up), direction\_view(C, B, front), progressToGoal(C, B)    0.0363782   \\
\item progressToGoal(A, B), rel\_with\_opp2(A, B, forward, down), rel\_with\_opp1(A, B, forward, up), rel\_with\_team(A, B, forward, up), direction\_view(A, B, front),     
next\_a(A, C), rel\_with\_opp1(C, B, forward, up), direction\_view(C, B, front), progressToGoal(C, B)    0.0355654   \\
\item progressToGoal(A, B), rel\_with\_opp2(A, B, forward, down), rel\_with\_opp1(A, B, forward, up), direction\_view(A, B, front),     
next\_a(A, C), pass(C, B, D), rel\_with\_opp2(C, D, forward, up), rel\_with\_opp1(C, D, forward, up),\\ direction\_view(C, D, left)    0.0348619   \\
\item progressToGoal(A, B), rel\_with\_opp2(A, B, forward, down),\\ rel\_with\_opp1(A, B, forward, up), rel\_with\_team(A, B, forward, up),     
next\_a(A, C), rel\_with\_opp1(C, B, forward, up), rel\_with\_team(C, B, forward, up), direction\_view(C, B, front), progressToGoal(C, B)    0.0348601   \\
\item progressToGoal(A, B), rel\_with\_opp2(A, B, forward, down), rel\_with\_opp1(A, B, forward, up), rel\_with\_team(A, B, forward, up), direction\_view(A, B, front),     
next\_a(A, C), rel\_with\_opp1(C, B, forward, up), rel\_with\_team(C, B, forward, up), progressToGoal(C, B)    0.0348601   \\
\item progressToGoal(A, B), rel\_with\_opp2(A, B, forward, down), rel\_with\_opp1(A, B, forward, up), rel\_with\_team(A, B, forward, up), direction\_view(A, B, front),     
next\_a(A, C), rel\_with\_opp1(C, B, forward, up), rel\_with\_team(C, B, forward, up), direction\_view(C, B, front), progressToGoal(C, B)    0.0348601   \\
\item progressToGoal(A, B), rel\_with\_opp2(A, B, forward, down), rel\_with\_opp1(A, B, forward, up), rel\_with\_team(A, B, forward, up),     
next\_a(A, C), pass(C, B, D), rel\_with\_opp2(C, D, forward, up), rel\_with\_opp1(C, D, forward, up), direction\_view(C, D, left)  0.0334488    \\
\item progressToGoal(A, B), rel\_with\_opp2(A, B, forward, down), rel\_with\_opp1(A, B, forward, up), rel\_with\_team(A, B, forward, up), direction\_view(A, B, front),     
next\_a(A, C), pass(C, B, D), rel\_with\_opp1(C, D, forward, up),\\ direction\_view(C, D, left)    0.0334488   \\
\item progressToGoal(A, B), rel\_with\_opp2(A, B, forward, down), rel\_with\_opp1(A, B, forward, up), rel\_with\_team(A, B, forward, up), direction\_view(A, B, front),     
next\_a(A, C), pass(C, B, D), rel\_with\_opp2(C, D, forward, up),\\ rel\_with\_opp1(C, D, forward, up)    0.0334488   \\
\item progressToGoal(A, B), rel\_with\_opp2(A, B, forward, down), rel\_with\_opp1(A, B, forward, up), rel\_with\_team(A, B, forward, up), direction\_view(A, B, front),     
next\_a(A, C), pass(C, B, D), rel\_with\_opp2(C, D, forward, up),\\ rel\_with\_opp1(C, D, forward, up), direction\_view(C, D, left)    0.0334488   \\
\item progressToGoal(A, B), rel\_with\_opp2(A, B, forward, down), rel\_with\_opp1(A, B, forward, up), direction\_view(A, B, front),     
next\_a(A, C), rel\_with\_opp1(C, B, forward, up), rel\_with\_team(C, B, forward, up), direction\_view(C, B, front), progressToGoal(C, B)   0.0323264    \\
\item progressToGoal(A, B), rel\_with\_opp1(A, B, forward, up), rel\_with\_team(A, B, forward, up), direction\_view(A, B, front),     
next\_a(A, C), rel\_with\_opp1(C, B, forward, up), rel\_with\_team(C, B, forward, up), direction\_view(C, B, front), progressToGoal(C, B)    0.0307072   \\
\item progressToGoal(A, B), rel\_with\_opp2(A, B, forward, down), rel\_with\_team(A, B, forward, up), direction\_view(A, B, front),     
next\_a(A, C), rel\_with\_opp1(C, B, forward, up), rel\_with\_team(C, B, forward, up), direction\_view(C, B, front), progressToGoal(C, B)    0.0303901   \\
\item progressToGoal(A, B), rel\_with\_opp1(A, B, forward, up), rel\_with\_team(A, B, forward, up), direction\_view(A, B, front),     
next\_a(A, C), pass(C, B, D),\\ rel\_with\_opp2(C, D, forward, up), rel\_with\_opp1(C, D, forward, up), direction\_view(C, D, left)    0.0300923   \\ 
\item progressToGoal(A, B), rel\_with\_opp1(A, B, forward, up), rel\_with\_team(A, B, forward, up), direction\_view(A, B, front),     
next\_a(A, C), rel\_with\_opp2(C, B, forward, down), rel\_with\_team(C, B, forward, up), direction\_view(C, B, front), progressToGoal(C, B)    0.0250424   \\
\item getball(A, B), rel\_with\_opp2(A, B, forward, up), rel\_with\_opp1(A, B, forward, up), direction\_view(A, B, front),     
next\_a(A, C), rel\_with\_opp1(C, B, forward, up), direction\_view(C, B, front), intercept(C, B)    0.0213223   \\
\item progressToGoal(A, B), rel\_with\_opp2(A, B, forward, down), rel\_with\_team(A, B, forward, up), direction\_view(A, B, front),     
next\_a(A, C), rel\_with\_opp2(C, B, forward, down), rel\_with\_team(C, B, forward, up), direction\_view(C, B, front), progressToGoal(C, B)    0.0201827   \\
\item getball(A, B), rel\_with\_opp2(A, B, forward, up), direction\_view(A, B, front),     
next\_a(A, C), rel\_with\_opp2(C, B, forward, up), rel\_with\_opp1(C, B, forward, up), direction\_view(C, B, front), intercept(C, B)    0.0197279   \\
\item getball(A, B), rel\_with\_opp2(A, B, forward, up), rel\_with\_opp1(A, B, forward, up),     
next\_a(A, C), rel\_with\_opp2(C, B, forward, up), rel\_with\_opp1(C, B, forward, up), direction\_view(C, B, front), intercept(C, B)   0.0197279    \\
\item getball(A, B), rel\_with\_opp2(A, B, forward, up), rel\_with\_opp1(A, B, forward, up), direction\_view(A, B, front),     
next\_a(A, C), rel\_with\_opp2(C, B, forward, up), rel\_with\_opp1(C, B, forward, up), direction\_view(C, B, front), intercept(C, B)    0.0197279   \\
\item getball(A, B), rel\_with\_opp1(A, B, forward, up), direction\_view(A, B, front),     
next\_a(A, C), rel\_with\_opp2(C, B, forward, up), rel\_with\_opp1(C, B, forward, up), direction\_view(C, B, front), intercept(C, B)    0.0194534   \\
\item getball(A, B), rel\_with\_opp2(A, B, forward, up), rel\_with\_opp1(A, B, forward, up), direction\_view(A, B, front),     
next\_a(A, C), rel\_with\_opp2(C, B, forward, up), rel\_with\_opp1(C, B, forward, up), intercept(C, B)   0.0194534    \\ 
\item getball(A, B), rel\_with\_opp2(A, B, forward, up), rel\_with\_opp1(A, B, forward, up), direction\_view(A, B, front),     
next\_a(A, C), rel\_with\_opp2(C, B, forward, up), direction\_view(C, B, front), intercept(C, B)   0.0188371    \\
\item progressToGoal(A, B), rel\_with\_team(A, B, forward, up), direction\_view(A, B, front),     
next\_a(A, C), pass(C, B, D), rel\_with\_opp2(C, D, forward, up),\\ rel\_with\_opp1(C, D, forward, up), rel\_with\_team(C, D, forward, down), direction\_view(C, D, left)    0.0188371  
\item getball(A, B), rel\_with\_opp2(A, B, forward, up), rel\_with\_opp1(A, B, forward, up), direction\_view(A, B, front),     
next\_a(A, C), rel\_with\_opp2(C, B, forward, up), rel\_with\_opp1(C, B, forward, up), direction\_view(C, B, front)    0.0076135   \\
\end{itemize}

\textbf{Some examples of the most interesting patterns mined belonging to the \textit{CBR} team.}

\begin{itemize}
\item getball(A, B), rel\_with\_team(A, B, behind, up), direction\_view(A, B, backwards),     
next\_a(A, C), pass(C, B, D), rel\_with\_opp1(C, D, forward, down), rel\_with\_team(C, D, forward, up), direction\_view(C, D, right)    0.0281427   \\ 
\item getball(A, B), rel\_with\_team(A, B, behind, up), direction\_view(A, B, backwards),     
next\_a(A, C), pass(C, B, D), rel\_with\_opp2(C, D, forward, down), rel\_with\_opp1(C, D, forward, down), direction\_view(C, D, right)    0.0257733   \\
\item getball(A, B), direction\_view(A, B, backwards),     
next\_a(A, C), pass(C, B, D), rel\_with\_opp2(C, D, forward, down), rel\_with\_opp1(C, D, forward, down),\\ rel\_with\_team(C, D, forward, up), direction\_view(C, D, right)    0.0245968   \\ 
\item getball(A, B), rel\_with\_team(A, B, behind, up),     
next\_a(A, C), pass(C, B, D), rel\_with\_opp2(C, D, forward, down), rel\_with\_opp1(C, D, forward, down), rel\_with\_team(C, D, forward, up), direction\_view(C, D, right)    0.0243266   \\ 
\item getball(A, B), rel\_with\_team(A, B, behind, up), direction\_view(A, B, backwards),     
next\_a(A, C), pass(C, B, D), rel\_with\_opp2(C, D, forward, down), rel\_with\_team(C, D, forward, up), direction\_view(C, D, right)    0.0243266   \\ 
\item getball(A, B), rel\_with\_team(A, B, behind, up), direction\_view(A, B, backwards),     
next\_a(A, C), pass(C, B, D), rel\_with\_opp2(C, D, forward, down), rel\_with\_opp1(C, D, forward, down), rel\_with\_team(C, D, forward, up)\\    0.0240568   \\ 
\item getball(A, B), rel\_with\_team(A, B, behind, up), direction\_view(A, B, backwards),     
next\_a(A, C), pass(C, B, D), rel\_with\_opp2(C, D, forward, down), rel\_with\_opp1(C, D, forward, down), rel\_with\_team(C, D, forward, up), direction\_view(C, D, right)   0.0240568    \\ 
\item getball(A, B), rel\_with\_team(A, B, forward, down), direction\_view(A, B, front),     
next\_a(A, C), pass(C, B, D), rel\_with\_opp2(C, D, forward, down), rel\_with\_opp1(C, D, forward, down), direction\_view(C, D, right)   0.0219150   \\  
\item getball(A, B), rel\_with\_opp1(A, B, forward, down), rel\_with\_team(A, B, forward, down), direction\_view(A, B, front),     
next\_a(A, C), pass(C, B, D),\\ rel\_with\_opp1(C, D, forward, down), direction\_view(C, D, right)    0.0211173   \\ 
\item getball(A, B), rel\_with\_opp2(A, B, forward, down), rel\_with\_opp1(A, B, forward, down), direction\_view(A, B, front),     
next\_a(A, C), pass(C, B, D),\\ rel\_with\_opp1(C, D, forward, down), direction\_view(C, D, right)    0.0203171   \\ 
\item getball(A, B), rel\_with\_opp2(A, B, forward, down), rel\_with\_opp1(A, B, forward, down), rel\_with\_team(A, B, forward, down),     
next\_a(A, C), pass(C, B, D), rel\_with\_opp2(C, D, forward, down), rel\_with\_opp1(C, D, forward, down)   0.0200586   \\ 
\item getball(A, B), rel\_with\_opp1(A, B, forward, down), direction\_view(A, B, front),     
next\_a(A, C), pass(C, B, D), rel\_with\_opp2(C, D, forward, down),\\ rel\_with\_opp1(C, D, forward, down), direction\_view(C, D, right)    0.0200527   \\ 
\item getball(A, B), rel\_with\_opp2(A, B, forward, down), rel\_with\_opp1(A, B, forward, down), direction\_view(A, B, front),     
next\_a(A, C), pass(C, B, D),\\ rel\_with\_opp2(C, D, forward, down), rel\_with\_opp1(C, D, forward, down)    0.0195564   \\ 
\item getball(A, B), rel\_with\_opp2(A, B, forward, down), rel\_with\_opp1(A, B, forward, down),     
next\_a(A, C),  pass(C, B, D), rel\_with\_opp2(C, D, forward, down), rel\_with\_opp1(C, D, forward, down), direction\_view(C, D, right)\\    0.0192615   \\ 
\item getball(A, B), rel\_with\_opp1(A, B, behind, up), rel\_with\_team(A, B, behind, up), direction\_view(A, B, backwards),     
next\_a(A, C), pass(C, B, D), rel\_with\_opp1(C, D, forward, down), direction\_view(C, D, right)    0.0185056   \\ 
\end{itemize}

\bibliographystyle{splncs}
\bibliography{biblio}

\end{document}